\documentclass[sigconf]{acmart}

\usepackage{caption}
\usepackage{tabularx}
\usepackage{multirow}
\usepackage{enumerate}
\usepackage{threeparttable} 
\usepackage{booktabs}
\usepackage{makecell}
\usepackage{algorithm}
\usepackage{algorithmic}
\AtBeginDocument{%
  \providecommand\BibTeX{{%
    \normalfont B\kern-0.5em{\scshape i\kern-0.25em b}\kern-0.8em\TeX}}}

\copyrightyear{2021}
\acmYear{2021}
\setcopyright{acmcopyright}\acmConference[MM '21]{Proceedings of the 29th ACM
International Conference on Multimedia}{October 20--24, 2021}{Virtual Event, China}
\acmBooktitle{Proceedings of the 29th ACM International Conference on Multimedia
(MM '21), October 20--24, 2021, Virtual Event, China}
\acmPrice{15.00}
\acmDOI{10.1145/3474085.3475421}
\acmISBN{978-1-4503-8651-7/21/10}

\acmSubmissionID{mfp1465}

\begin{document}
\fancyhead{}
\title{Learning Fine-Grained Motion Embedding \\ for Landscape Animation}

\author{Hongwei Xue}
\authornote{This work was performed when Hongwei Xue was visiting Microsoft Research as a research intern.}
\email{gh051120@mail.ustc.edu.cn}
\affiliation{%
  \institution{University of Science and Technology of China}
  \city{Hefei}
  \country{China}
}

\author{Bei Liu}
\email{bei.liu@microsoft.com}
\affiliation{%
  \institution{Microsoft Research}
  \city{Beijing}
  \country{China}
}

\author{Huan Yang}
\email{huayan@microsoft.com}
\affiliation{%
  \institution{Microsoft Research}
  \city{Beijing}
  \country{China}
}

\author{Jianlong Fu}
\email{jianf@microsoft.com}
\affiliation{%
  \institution{Microsoft Research}
  \city{Beijing}
  \country{China}
}

\author{Houqiang Li}
\email{lihq@ustc.edu.cn}
\affiliation{%
  \institution{University of Science and Technology of China}
  \city{Hefei}
  \country{China}
}

\author{Jiebo Luo}
\email{jluo@cs.rochester.edu}
\affiliation{%
  \institution{University of Rochester}
  \city{Rochester}
  \country{NY}
}

\renewcommand{\shortauthors}{Xue and Liu, et al.}

\begin{abstract}
  In this paper we focus on landscape animation, which aims to generate time-lapse videos from a single landscape image. Motion is crucial for landscape animation as it determines how objects move in videos. Existing methods are able to generate appealing videos by learning motion from real time-lapse videos. However, current methods suffer from inaccurate motion generation, which leads to unrealistic video results. To tackle this problem, we propose a model named \textbf{FGLA} to generate high-quality and realistic videos by learning \textbf{F}ine-\textbf{G}rained motion embedding for \textbf{L}andscape \textbf{A}nimation. Our model consists of two parts: (1) a motion encoder which embeds time-lapse motion in a fine-grained way. (2) a motion generator which generates realistic motion to animate input images. To train and evaluate on diverse time-lapse videos, we build the largest high-resolution \textbf{Time-lapse} video dataset with \textbf{D}iverse scenes, namely \textbf{Time-lapse-D}, which includes 16,874 video clips with over 10 million frames. Quantitative and qualitative experimental results demonstrate the superiority of our method. In particular, our method achieves relative improvements by 19\% on LIPIS and 5.6\% on FVD compared with state-of-the-art methods on our dataset. A user study carried out with 700 human subjects shows that our approach visually outperforms existing methods by a large margin.
\end{abstract}





\maketitle

\begin{figure}[!tp]
\centering
\includegraphics[width=\linewidth]{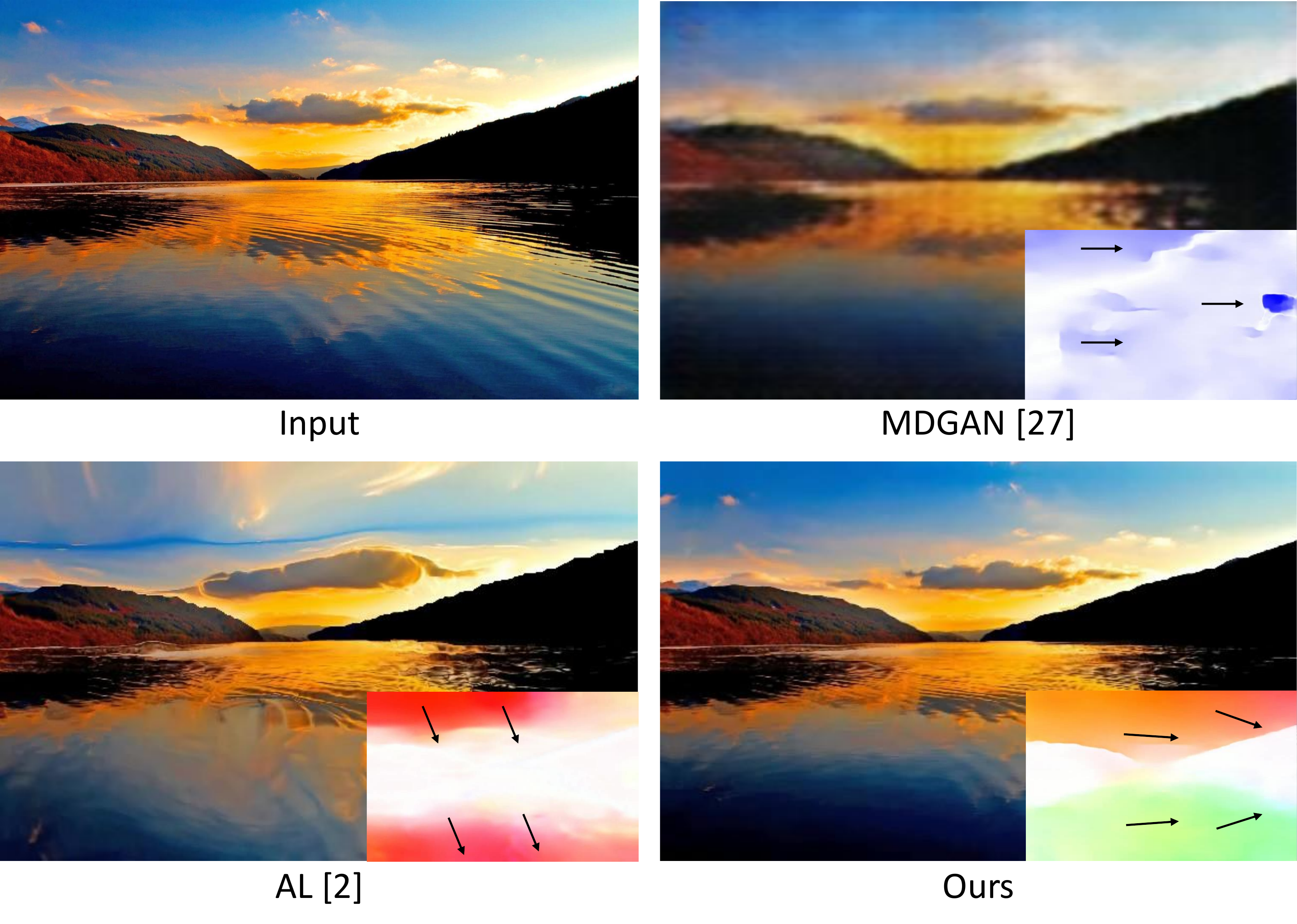}
\vspace{-7mm}
\caption{Methods implicitly learning motion (e.g., MDGAN \cite{xiong2018learning}) tend to generate videos with subtle and irrational motion. Explicit motion learning by direct encoding optical flows (e.g., AL \cite{endo2019animating}) can alleviate this but sometimes generates inaccurate motion. We visualize optical flows to show the motion. Different colors of visualized optical flows represent different motion patterns. The motion direction is roughly marked by black arrows. [Best viewed in color.]} \label{figure:intro}
\vspace{-5mm}
\end{figure}

\vspace{-2mm}
\section{Introduction} \label{intro}
Image animation aims to generate a time-lapse video from a still image in an automatic way. This topic has drawn great attention in recent years \cite{hu2018video,zhao2018learning,zhou2018image2gif}, as video representation has become an evolving trend in content consumption that makes people more engaged. More specifically, landscape animation \cite{endo2019animating,xiong2018learning,zhang2020dtvnet} can benefit a broad range of applications, such as video footage production, social engagement boosting, virtual background animation for online video conferences. Thus, we mainly focus on landscape animation in this paper.
For landscape videos, motion is crucial as it determines how objects move in the video (e.g., moving clouds, flowing water). However, motion learning is quite difficult due to motion's uncertainty and complexity. Uncertainty means it is difficult to predict the future from one image. For example, clouds in the sky might move left, right, forward, or backward in next frames. Thus, the model requires capabilities of learning diverse motion. Meanwhile, motion in a landscape video is complex as objects usually have different motion patterns and each pattern should be well aligned to the object. Therefore, although motion learning is crucial, it's challenging to learn diverse and realistic motion.

Existing works on landscape animation model motion in an implicit or explicit way. GAN-based methods use 3D-CNNs to implicitly learn motion from real videos \cite{xiong2018learning}. This implicit learning usually results in subtle and irrational motion, which does not correspond to scene's layout, such as the result of MDGAN \cite{xiong2018learning} in Figure \ref{figure:intro} (right-top). This is because the spatial-temporal domain
of videos is too large for generative models to learn. Besides, the lack of motion embedding leads to a lack of diverse video generation capabilities, which limits the practicality of these methods. To solve these problems, some works use an encoder-decoder paradigm to explicitly learn motion by using a convolutional encoder to embed estimated optical flows into a latent space \cite{zhang2020dtvnet,endo2019animating}. Relying on optical flows as the representation of motion, these works are able to well learn motion then generate vivid videos. Also, the capacity of generating diverse videos can be supported by sampling from motion embedding latent space.

Although current motion embedding can well handle the uncertainty of future frames, it still suffers from motion's complexity, especially when an image involves complex scenes. Compared to images, optical flows lack of semantics that can be learned and distinguished by convolutional networks as it is only the derivative of pixel coordinates. Thus, the generated videos often have inaccurate motion, including inaccurate alignment with scene's layout and unified motion among global frame (see Figure \ref{figure:intro}, AL \cite{endo2019animating}). Inaccurate motion may cause the objects in videos to move unrealistically and even distorted. 

To better learn motion for a more realistic landscape animation, we propose a novel fine-grained motion embedding learning approach. Our approach is able to embed motion information well then generate accurate and realistic motion. To achieve this, our model consists of a fine-grained motion encoder and a motion generator. The fine-grained motion encoder decomposes the motion space into a fine-grained latent space guided by semantics. To be more specific, our fine-grained motion encoder is based on partial convolution by which motion space can be well decomposed without interference with each other. Moreover, the scaling factor of each partial convolution ensures that the motion information of small objects or textures in the frame will not be ignored. Then a latent map and the input image are spatially aligned then fed into the motion generator to generate motion for animation. The output video is generated frame-by-frame by warping the input image according to the generated motion.


Existing works mostly focus on sky time-lapse videos. To train and evaluate on diverse videos, we introduce a new dataset, namely \textbf{Time-lapse-D}, which consists of 16,874 high-resolution \textbf{Time-lapse} videos of \textbf{D}iverse scene with category annotations. The videos in our dataset have over 10 million high-resolution frames in total, of which over 40\% are $1280 \times 720$ and over 30\% are $1920 \times 1080$ resolution. To the best of our knowledge, this is the largest time-lapse video dataset so far, which can provide large-scale and high-quality data support for image animation tasks. 

Quantitative and qualitative experiments on our dataset demonstrate the superiority of our proposed method. We also conduct a user study to evaluate the comparison of our method with existing works and the confusion ratio with the real video. These experiments demonstrate that our method outperforms existing works by a large margin. Besides, our fine-grained learning makes it possible to control different objects in frames with different motion patterns.

Our key contributions are as follows:
\begin{enumerate}[1]
\item We propose a novel fine-grained motion embedding method for the landscape animation task to generate high-quality video with realistic motion.
\item We build the largest time-lapse video dataset (Time-lapse-D) which includes diverse natural scenes with over 10 million frames.
\item We conduct extensive experiments, including subjective and objective comparison experiments and qualitative experiments. Our method achieves relative improvements by 19\% on LIPIS and 5.6\% on FVD compared with state-of-the-art methods on our dataset.
\end{enumerate}

\section{Related Work}

\begin{figure*}[t]
\centering
\includegraphics[width=\linewidth]{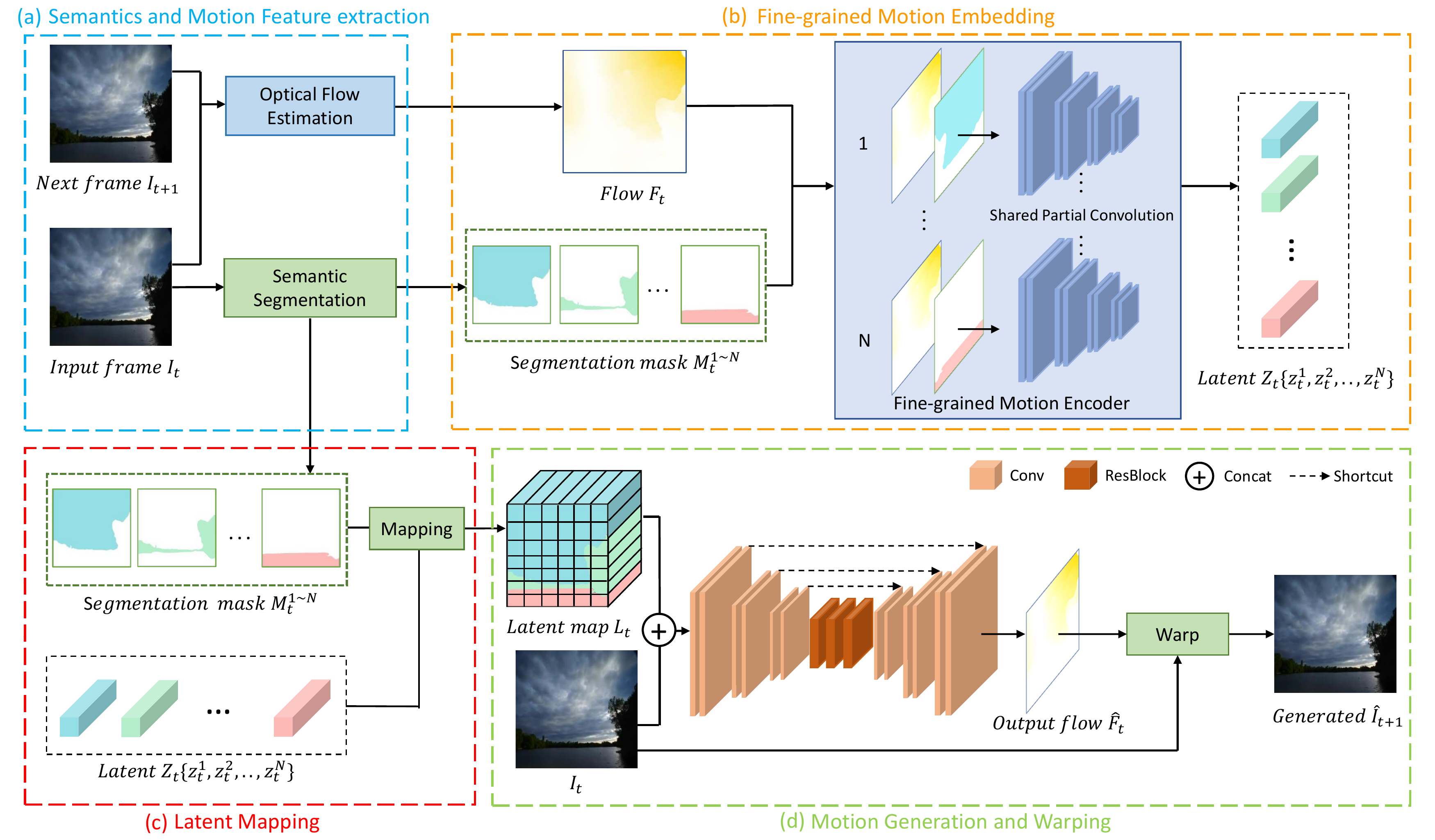}
\vspace{-7mm}
\caption{The overall pipeline of FGLA. During motion embedding phase, (a) we first extract semantics and motion features by a semantic segmentation module and an optical flow estimation module respectively; (b) a fine-grained motion encoder takes the optical flow and segmentation masks of the input image then generates a latent set. Different colors in latent set $Z_t$ represent different semantics. For frame generation, (c) the semantic segmentation masks and the latent set are reorganized into a latent map which is fed into a generator with the input frame; (d) the generator generates an optical flow for the motion to warp the input frame to obtain the generated frame. Different colors of visualized optical flow represent different directions and the higher saturation means higher speed and vice versa. [Best viewed in color.]} \label{figure:framework}
\end{figure*}

In this section, we briefly review related works of general video generation from a still image. Then we specifically summarize related works on landscape animation.

\noindent\textbf{Video Generation from Still Image.}
Traditional non-learning-based approaches using examples can reproduce realistic motion and content without complex parameters by directly transferring reference videos \cite{okabe2009animating,okabe2011creating,okabe2018animating,prashnani2017phase,shih2013data}. These approaches heavily depend on large datasets and huge computational resources. Also, existing techniques often impose heavy manual processes for specifying, e.g., alpha mattes, flow fields and regions for fluid. 

The past few years have witnessed dramatic advances in learning-based approaches. Some Generative Adversarial Networks (GAN) \cite{goodfellow2014generative} based models \cite{tulyakov2018mocogan,zhao2018learning,zhou2018image2gif} and conditional variational auto-encoders (CVAE) based models \cite{li2018flow,pan2019video} automatically generate videos conditioned on the content of input images. Among them, explicitly motion learning is achieved by using sparse trajectories \cite{hao2018controllable}, human poses \cite{zhao2018learning}, or optical flows \cite{li2018flow}. In this paper, we adapt optical flows as motion representation as flow estimation approach has been well developed.


\noindent\textbf{Landscape Animation.}
Landscape animation targets to generate a time-lapse video from a landscape image. This task can be decomposed into two subtasks: color animation and motion animation \cite{endo2019animating}. Color animation is animating the color of a landscape at different hours in one day (e.g., time-varying colors in the sky) \cite{cheng2020time,nam2019end}. Motion animation is animating the scene at a smaller time scale (e.g., moving clouds) \cite{endo2019animating,logacheva2020deeplandscape,xiong2018learning,zhang2020dtvnet}. As the data required by these two subtasks are at different time scales, existing works mostly focus on one subtask. We focus on and refer to motion animation in the rest of this paper. It is worth noting that we only refer to motion animation part when mention AL \cite{endo2019animating} as it independently trains two models on different datasets to solve these two subtasks.


MDGAN \cite{xiong2018learning} learns the whole spatial-temporal space of videos and directly generates all video frames. MDGAN presents a 3D-CNN based two-stage approach which generates videos of realistic content for each frame in the first stage and then refines the generated video from the first stage.  Although MDGAN enforces generated videos to be closer to real
videos with regard to motion dynamics in the second stage, it's too hard for 3D-CNN to model the whole spatial-temporal domain, resulting in subtle and irrational motion. Besides, color distortion and blur are observed on their results. DTVNet \cite{zhang2020dtvnet} and AL \cite{endo2019animating} explicitly learn the motion by motion embedding. DTVNet introduces optical flow into GAN and consists of two submodules. An optical flow encoder embeds a sequence of optical flow maps to a normalized motion vector. A dynamic video generator generates all frames from the motion vector and the input image. DTVNet alleviates color distortion but still suffers from the blur problem, and also tends to generate videos with subtle motion. AL \cite{endo2019animating} 
uses optical flows in both embedding and generating stages. An encoder-decoder paradigm is used to encode motion into a latent space and then generates optical flow for warping. A video is generated frame-by-frame by warping the input image. Results generated by AL have high-quality frames without blur or color distortion. However, the insufficiency of motion modeling often leads to inaccurate and global monotonic motion thus fails in complex scenes. In this paper, we improve motion modeling by embedding fine-grained motion guided by semantics. The inaccurate and global motion problem is well alleviated by our approach.

Some methods model motion by sampling from pre-defined transformations \cite{logacheva2020deeplandscape}. DeepLandscape \cite{logacheva2020deeplandscape} extends StyleGAN model \cite{karras2019style} by augmenting it with learning pairwise consistency of frames in videos. Their generated video frames are sometimes discontinuous due to their method does not learn full temporal dynamics of videos. As a result, They can generate appealing videos of time-lapse clouds but often fails on other scenes.

\section{Approach}\label{sec:approach}

In this section, we will introduce the proposed FGLA model from aspects of model structure, network training and fine-grained controllable generation. Figure \ref{figure:framework} shows the whole pipeline of our approach. We first extract semantics by image segmentation and motion features in optical flow. During motion embedding, the fine-grained motion encoder takes a backward optical flow $F_t$ and semantic segmentation masks $M_t$ of the input frame $I_t$ and then generates a latent set $Z_t$. Each variable in the set is corresponding to a unique semantic. Then the latent set is organized as a latent map $L_t$ fed into a motion generator which takes the input frame and generates an optical flow $\hat{F}_t$ to warp the input frame. In this section we will introduce the details.

\subsection{Fine-grained Motion Embedding}

To extract semantic and motion features of an image, we use ADE20k-pretrained \cite{zhou2019semantic} segmentation network to obtain a set of segmentation binary masks ${M_t^1,M_t^2,...,M_t^N}$ and use FlowNet 2.0 \cite{ilg2017flownet} to estimate optical flow $F_t$ of an image pair $(I_t, I_{t+1})$.

One landscape video usually have unique motion patterns for different semantics. To independently encode the motion information of each semantic part, we adopt partial convolution \cite{liu2018image} as our extractor. In each layer of partial convolution, the filter weights $\mathbf{W}$ and the corresponding bias $b$ are learnable parameters. $\mathbf{X}$ and $\mathbf{M}$ denote the input feature maps and the corresponding binary mask in each partial convolution layer. The partial convolution is expressed as:
\setlength\abovedisplayskip{1pt}
\begin{equation}
x^{\prime}=\left\{\begin{array}{ll}
\mathbf{W}^{T}(\mathbf{X} \odot \mathbf{M}) \frac{\operatorname{sum}(\mathbf{1})}{\operatorname{sum}(\mathbf{M})}+b, & \text { if } \operatorname{sum}(\mathbf{M})>0, \\
0, & \text { otherwise, }
\end{array}\right.
\end{equation}
where $\odot$ denotes element-wise multiplication and $\operatorname{sum}$ denotes element summation. We use the same way of updating the binary mask as in \cite{liu2018image}:
\begin{equation}
m^{\prime}=\left\{\begin{array}{ll}
1, & \text { if } \operatorname{sum}(\mathbf{M})>0, \\
0, & \text { otherwise, }
\end{array}\right.
\end{equation}

Through partial convolutional computing, the output semantic latent variables depend only on their corresponding semantics. The scaling factor $\operatorname{sum}(\mathbf{1})/\operatorname{sum}(\mathbf{M})$ adjusts for the differences in proportion of different semantics. 

We denote the fully partial convolution network $\mathcal{E}$ in the encoder. With each binary mask $M_t^i$ in $t$-th frame of $i$-th semantics, we feed the flow $F_t$ into the encoder $\mathcal{E}$ to generate latent variable $z_t^i$ for each semantics separately as follows: 
\begin{equation}\label{entropy}
z_t^i = \mathcal{E}(F_t, M_t^i).
\end{equation}
As a result, we get a set of latent variables and each variable $z_t^i$ models the motion information of a unique semantic:
\begin{equation}
    Z_t = [z_t^{1}, z_t^{2}, ..., z_t^{N}].
\end{equation}

\subsection{Latent Mapping and Motion Generation}
The main challenge of landscape animation is to drive the given image to transform realistically over time and the content will not change too much. There are two advantages of the flow-warping strategy: realistic and high-fidelity. First, every pixel of generated frames is transformed from pixels of original images, making results realistic. Second, an optical flow is smooth in a small neighborhood. As a result, an optical flow can be interpolated to high resolution. Therefore our model can directly generate high-fidelity landscape videos, without using interpolation or super-resolution techniques to increase frames' size. For above reasons, our generator generates frames based on flow-warping manner: we first generate optical flow for the motion, and then warp the input image with optical flow for next frame.

The learned semantic latent variables are supposed to represent motion of each semantics without the spatial information of videos. Thus, we propose a latent mapping module to combine the image content and its corresponding latent variables into one latent map. Given the given landscape image $I_t$ and the obtained latent $Z_t$, we reorganize the latent set as the latent map $L_t$ by filling the semantic mask $M_t$ with the corresponding latent as follows:
\begin{equation}
    L_t(x, y) = z_k  \quad when \quad  M_t^k(x, y) = 1, 
\end{equation}
where $(x,y)$ is coordinate. Then we concatenate $L_t$ with $I_t$ and feed them into the motion generator to generate optical flow:
\begin{equation}\label{entropy}
\hat{F}_t = \mathcal{P}(L_t \oplus I_t).
\end{equation}
Similar to \cite{endo2019animating,pan2019video}, we use skip-connections in motion generator. This U-Net \cite{isola2017image} like architecture generates an optical flow that spatially corresponds well to the image.

Using $tanh$, our generator infers an optical flow $\hat{F}_t$ which value ranges from -1 to 1. Similar to \cite{endo2019animating}, we divide the flow by a constant $c > 1$ since the objects do not move significantly in a single timestep. While we choose a larger $c$ to generate optical flow with a larger amplitude range. In this paper, we set $c$ to 32. 

With the generated optical flow $\hat{F}_t$, the input image $I_t$ is warped to generate next frame $\hat{I}_{t+1}$ as follows:
\begin{equation}
\hat{I}_{t+1} = \operatorname{Warp}(I_t, \hat{F}_t).
\end{equation}

\subsection{Network Training}

\setlength{\textfloatsep}{2pt}
\begin{algorithm}[!t]
    \renewcommand{\algorithmicrequire}{\textbf{Input:}}
    \renewcommand{\algorithmicensure}{\textbf{Output:}}
    \caption{Fine-grained Controllable Landscape Animation}
    \label{alg:infer}
    \begin{algorithmic}[1]
        \REQUIRE An input image $I_0$, A set of reference videos $R_1, R_2, ..., R_L$.
        \ENSURE A generated video $V=\{\hat{I}_{0}, \hat{I}_{1}, ..., \hat{I}_{T}\}$.
        \STATE Load a motion encoder $\mathcal{E}$, a motion generator $\mathcal{P}$;
        \STATE Let $\hat{I}_{0}=I_0$, $t = 0$;
        \WHILE {$t < T$}
        \STATE Compute segmentation masks ${M_t^1,M_t^2,...,M_t^N}$ for $\hat{I}_{t}$;
        \STATE Compute optical flows $F_t^{k}$ and segmentation masks ${M_t^{1,k},M_t^{2,k},...,M_t^{N,k}}$ for each reference video $R_k$;
        \STATE Compute $z_t^{i,k} = \mathcal{E}(F_t^k, M_t^{i,k})$, $i \in [1, N]$ is the $i$th semantic class. Then get $Z_t^k = [z_t^{1,k}, z_t^{2,k}, ..., z_t^{N,k}]$ for each reference video $R_k$;
        \STATE Controllably get $Z_t = [z_t^{1, r_1}, z_t^{2, r_2}, ..., z_t^{N, r_N}]$, where $z_t^{i, r_i}$ is the $i$th semantic latent variable of the $r_i$th reference video, $r_i \in [1, L]$.
        \STATE Fill a latent map $L_t(x, y) = Z_t[i]  \quad when \quad  M_t^i(x, y) = 1$ for each pixel $(x, y)$ in latent map;
        \STATE Compute generated optical flow $\hat{F}_t = \mathcal{P}(L_t \oplus \hat{I}_{t})$, where $\oplus$ is concatenation;
        \STATE Accumulate flow $\hat{F}_{0 \rightarrow t} = \hat{F}_t + \hat{F}_{0 \rightarrow t-1} \quad if \quad t>0 \quad else \quad \hat{F}_{0}$
        \STATE Warp the input image $I_0$ to generate frame $\hat{I}_{t+1} = \operatorname{Warp}(I_0, \hat{F}_{0 \rightarrow t})$
        \STATE t = t + 1;
        \ENDWHILE
    \end{algorithmic}
\end{algorithm}

\begin{figure*}[t]
\centering
\includegraphics[width=0.8\paperwidth]{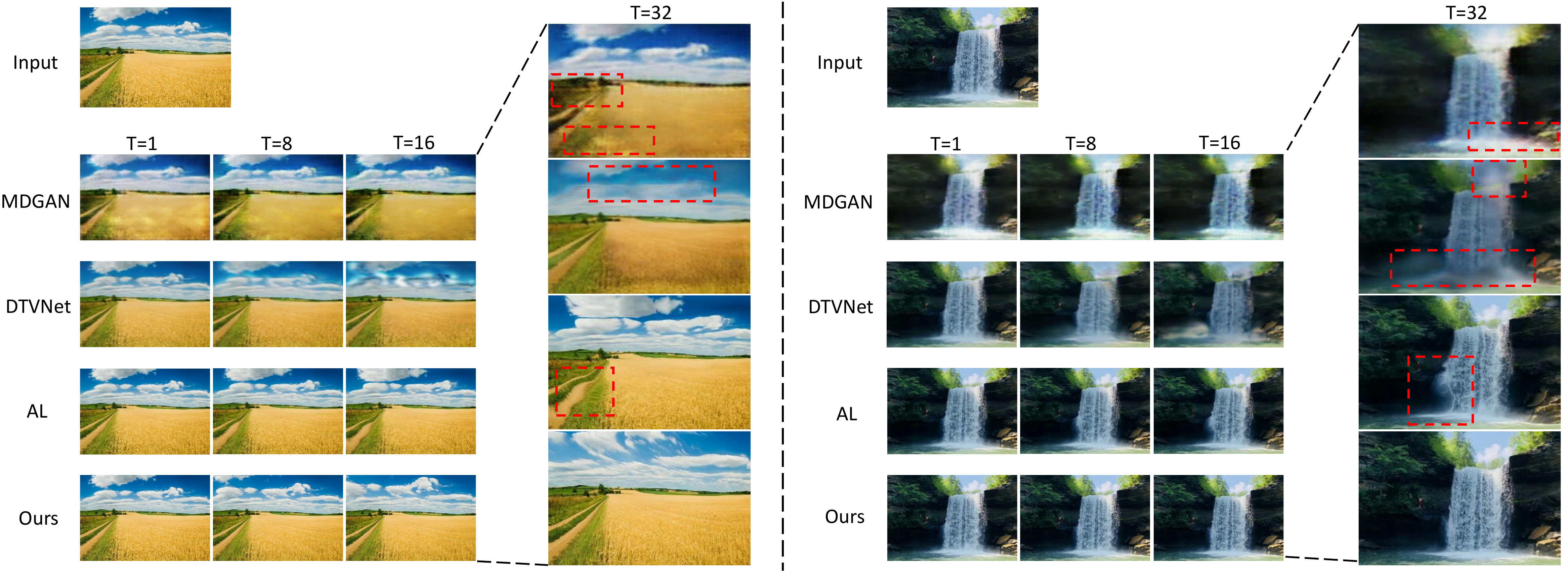}
\vspace{-4mm}
\caption{Comparisons of frames sampled from videos generated by MDGAN \cite{xiong2018learning}, DTVNet \cite{zhang2020dtvnet}, AL \cite{endo2019animating} and ours. To view details, we zoom in the last frames. We highlight some defects with red rectangles. [Best viewed in color.]} \label{figure:qualitative}
\vspace{-4mm}
\end{figure*}

For supervised training of our landscape animation model, we require a ground-truth video of each input image. However, this kind of ground-truth videos are very difficult to be obtained. For this reason, we utilize a self-supervised training mechanism by using the horizontally flipped version as the ground-truth video similar to \cite{cheng2020time}. Flipping can prevent our model from learning the spatial correlation between input images and ground-truth videos. To keep the value of the optical flow constant in each pixel, we conduct the flip operation after estimating the optical flow on the original image.

 
The training objectives of the landscape animation framework consists of three parts: frame loss, flow loss, and edge-preserving smoothing loss. We adopt a L1 loss to compute distance between the generated output frame $\hat{I}_{t+1}$ and the ground-truth image $I_{t+1}$:
\begin{equation}
\mathcal{L}_{frame} = \left\|\hat{I}_{t+1} - I_{t+1}\right\|_{1}. 
\end{equation}
As the warping operation is sampling and interpolating pixels on the original image, the frame loss has very limited effect in areas with small color changes or some color patterns. We add an L1 loss to measure the difference of generated optical flow $\hat{F}$ and the ground-truth optical flow $F$ in each timestep:
\begin{equation}
\mathcal{L}_{flow} = \left\|\hat{F}_t - F_t\right\|_{1}. 
\end{equation}
Similar to \cite{endo2019animating}, we adopt a weighted total variation loss applied to the output flow field for the purpose of edge-preserving smoothing:
\begin{equation}
\mathcal{L}_{tv}=\sum_{\mathbf{p}, \mathbf{q} \in N(\mathbf{p})} w\left(\hat{I}_{t+1}(\mathbf{p}), \hat{I}_{t+1}(\mathbf{q})\right)\left\|\hat{F}_t(\mathbf{p})-\hat{F}_t(\mathbf{q})\right\|_{1},
\end{equation}
\begin{equation}
w(\mathbf{x}, \mathbf{y})=\exp \left(-\frac{\|\mathbf{x}-\mathbf{y}\|_{1}}{\sigma}\right),
\end{equation}
where $N(\mathbf{p})$ indicates the right and above neighbors of $\mathbf{p}$. This loss function makes the model tend to output close optical flows in areas where the pixel value changes little. The final objective of our model is the combination of the above three losses:
\begin{equation}
\mathcal{L}_{total}= \mathcal{L}_{frame} + \alpha*\mathcal{L}_{flow} + \beta*\mathcal{L}_{tv},
\end{equation}
where $\alpha$ and $\beta$ are the weights of $\mathcal{L}_{flow}$ and $\mathcal{L}_{tv}$ respectively.

\subsection{Fine-grained Controllable Animation}
Instead of randomly sampling from a latent codebook or latent space, our approach makes it possible to generate videos in a fine-grained controllable manner. Each variable in the latent set represents one unique semantics. This kind of latent set makes it possible to transfer semantic motion patterns from one or multiple reference videos. We can simply get latent $Z_t$ with different semantics from multiple references $R_1, R_2, ..., R_L$:
\setlength\abovedisplayskip{1pt}
\setlength{\belowdisplayskip}{1pt}
\begin{equation}
    Z_t = [z_t^{1, r_1}, z_t^{2, r_2}, ..., z_t^{N, r_N}], 
\end{equation}
where $z_t^{i, r_i}$ is the $i$th semantic latent variable of the $r_i$th reference video, $r_i \in [1, L]$. The whole inference procedure is shown in Algorithm \ref{alg:infer}.

During inference, similar to \cite{endo2019animating}, we recurrently use the generated frame $\hat{I}_{t}$ as the input of next motion generation while we warp optical flows sequentially to accumulate flows so that we can reconstruct each output frame from the first frame. Through this design, we can avoid generating gradually blurry frames due to error accumulation in multiple sampling. 

\section{Experiments}\label{sec:experiments}
\subsection{Dataset}

\begin{table}[]
\small
\centering
\begin{threeparttable}
    \caption{Comparison between Sky Time-lapse \cite{xiong2018learning} and our proposed dataset.}
    \begin{tabularx}{\linewidth}{l c c}
         \toprule
         \hfill & Sky Time-lapse \cite{xiong2018learning} & Time-lapse-D\\
         \midrule
         Clips             & 2049 & 16874    \\ 
         Frames            & $1.2 \times 10^{6}$  & $1.0 \times 10^{7}$    \\ 
         Resolution      & $640\times360$ & \makecell[c]{40\% $1280\times720$\\30\% $1920\times1080$} \\ 
         Scene Categories   & 2 &  13  \\
         \bottomrule
    \end{tabularx}
    \label{tab:dataset}
\end{threeparttable}
\end{table}

    

\begin{table*}[htbp]
\centering
\begin{threeparttable}
    \caption{Comparison with existing works and ablation study on Time-lapse-D. The $\uparrow$ indicates that the larger the value, the better the model performance, and vice versa for $\downarrow$. For PSNR and SSIM, we mask out dynamic region pixels.}
    
    \setlength{\tabcolsep}{5mm}
    \begin{tabular}{lccccc}
         \toprule
         Method  & Masked PSNR $\uparrow$ & Masked SSIM $\uparrow$ & FID $\downarrow$ & LPIPS $\downarrow$ & FVD $\downarrow$\\
         \midrule
         MDGAN \cite{xiong2018learning}  & 27.85  &  0.841 & 60.38  & 0.225 & 1177.75  \\
         DTVNet \cite{zhang2020dtvnet}   & 31.41  &  0.904 & 59.37  & 0.165 & 1007.81 \\
         AL \cite{endo2019animating}  & 42.77  & 0.981  & 59.08  & 0.026 & 831.46 \\
         Ours w/o LM  & 41.38  &  0.981 & 59.27  & 0.028 & 901.62  \\
         Ours w/o flip  & 44.23  &  0.980 & 58.78  & 0.022 & 803.26  \\
         Ours w/ $\mathcal{L}_{flow}$ & 42.01  &  0.979 & 58.79  & 0.025 & 799.71 \\
         Ours w/ $\mathcal{L}_{frame}$ & \textbf{45.61}  & \textbf{0.984}  & 59.03  & 0.025 & 802.42 \\
         Ours w/o $\mathcal{L}_{tv}$ & 44.99  & 0.980  & 58.95  & 0.024 & 787.33 \\
         Ours &  45.53 & 0.982  &  \textbf{58.77} &  \textbf{0.021} & \textbf{784.99} \\
         \bottomrule
    \end{tabular}
    \label{tab:comparison1}
\end{threeparttable}
\end{table*}

Time-lapse videos are usually used for the experiments of landscape animation. One recent dataset that is used by \cite{endo2019animating,zhang2020dtvnet} is Sky Time-lapse dataset \cite{xiong2018learning},  which contains dynamic sky scenes, such as the cloudy sky with moving clouds and the starry sky with moving stars. There are 1,825 training video clips and 224 testing video clips. Note that by video clips we refer to real video clips instead of divided fixed-length clips. The original size of each frame is $3 \times 640 \times 360$. This dataset contains a small number of video clips and is limited to sky scenes. To better model the motion of time-lapse videos, we build a new large dataset by collecting more than 10,000 high-resolution time-lapse videos from Flickr. By PySceneDetect\footnote{https://pyscenedetect.readthedocs.io}, these videos are automatically cut into video clips based on the scene variety. 
We utilize crowdsourcing to manually filter out clips that have obvious camera movement or quick zoom-in/zoom-out. We then have the remaining video clips annotated with categories of different dynamic scenes, such as sky, water and hayfield.

The dataset is named as \textbf{Time-lapse-D}, in which ``D" indicates diverse scenes. The statistics of Time-lapse-D are shown in Table \ref{tab:dataset}. Time-lapse-D contains 16,874 video clips with diverse scene categories in total, includes 14,418 clips for training and 2,456 clips for testing. It is about 8 times larger than Sky Time-lapse dataset \cite{xiong2018learning}. The videos in the test set are selected from different scene categories in the same proportion, to ensure that the test set contains all categories of scenes. Each video clip contains 607 frames on average. Most of the clips are with high resolution, with over 40\% are $1280 \times 720$ and over 30\% are $1920 \times 1080$. The dynamic patterns in our dataset are much more diverse, including common categories (i.e., cloud, star, water, waterfall, hayfield, traffic, smoke, tree, sun, moon, aurora, fire, fog).  

This is the largest high-resolution time-lapse video dataset with the most diverse categories. In this paper, we mainly focus on five moving scenes that are feasible for flow-warping (i.e., sky, tree, grass, water and waterfall). Even though some categories (e.g., traffic and smoke) are not our targets in this work, this dataset provides high-quality and diverse video data to facilitate the research of video prediction, classification, generation and dynamic texture synthesis.

\subsection{Implementation Details}

\begin{table}[]
\centering
\begin{threeparttable}
    \caption{User study results. Our approach outperforms other approaches by a large margin.}
    \begin{tabular}{ll}
         \toprule
         Comparison Methods & User Preference \\
         \midrule
         MDGAN vs.\ Ours   & 4 vs.\ 96     \\ 
         DTVNet vs.\ Ours  & 6 vs.\ 94   \\ 
         AL vs.\ Ours      & 22 vs.\ 78    \\ \hline
         MDGAN vs.\ GT   & 2 vs.\ 98     \\ 
         DTVNet vs.\ GT  & 4 vs.\ 96   \\ 
         AL vs.\ GT      & 13 vs.\ 87   \\
         Ours vs.\ GT     & \textbf{31 vs.\ 69 }  \\
         \bottomrule
    \end{tabular}
    \label{tab:userstudy}
\end{threeparttable}
\end{table}

In this work, we use FlowNet 2.0 \cite{ilg2017flownet} as our optical flow estimator. The FlowNet 2.0 model takes two $3 \times 256 \times 256$ frames as input and estimates a $2 \times 256 \times 256$ backward optical flow as output. 
We use HRNetV2 \cite{sun2019high} as our segmentation model and fix the resolution of all outputs to $256 \times 256$. As we focus on five different types of scenes, we aggregate the categories of the data into six groups: sky, tree, grass, water, waterfall and others.
We horizontally flip and downsample the flow and semantic masks, then feed them into the fine-grained motion encoder. The encoder takes a $2 \times 128 \times 128$ flow and a $6 \times 128 \times 128$ mask as input and outputs a $6 \times 2$ latent matrix. As described in Section \ref{sec:approach}, we get a $2 \times 256 \times 256$ latent map and feed it into the generator. The motion generator takes the latent map and a $3 \times 256 \times 256$ frame as input to generate a $2 \times 256 \times 256$ optical flow.  During training stage, we use Adam \cite{kingma2014adam} optimizer for all modules with learning rate of $1e^{-4}$. Similar to \cite{endo2019animating}, Instance Normalization \cite{ulyanov2016instance} followed by Leaky ReLU \cite{xu2015empirical}with slope of 0.1 is used in all the convolution layers in both encoder and generator. We empirically set $\sigma=0.1$ for edge-preserving smoothing loss. The weights of the frame reconstruction loss, flow reconstruction loss and smoothing loss $\alpha$ and $\beta$ are set to 0.5 and 5 respectively.

To train our model, we first sample one from every four frames for each clip to avoid learning motion that is too subtle. Then we resize the frames into square images of size $256 \times 256$. Before feeding the frames to the model, we normalize the color values to $[-1, 1]$. No additional pre-processing is required. To avoid training biases by long video clips, we train each pair of frames sampled randomly for each video clip in each epoch. Under the above settings, our model converges in 500 epochs.

\subsection{Comparison with State-of-the-art methods}

\begin{table*}[htbp]
\centering
\addtolength{\tabcolsep}{8pt}
\begin{threeparttable}
    \caption{Comparison with existing works and ablation study on Sky Time-lapse \cite{xiong2018learning}. The $\uparrow$ indicates that the larger the value, the better the model performance, and vice versa for $\downarrow$. For PSNR and SSIM, we mask out dynamic region pixels.}
    \setlength{\tabcolsep}{5mm}
    \begin{tabular}{lccccc}
         \toprule
         Method  & Masked PSNR $\uparrow$ & Masked SSIM $\uparrow$ & FID $\downarrow$ & LPIPS $\downarrow$ & FVD $\downarrow$\\
         \midrule
          
         MDGAN \cite{xiong2018learning}  & 31.43  &  0.911 & 67.32  & 0.186 & 1086.66  \\
         DTVNet \cite{zhang2020dtvnet}   & 35.80  &  0.950 & 67.34  & 0.149 & 774.85 \\
         AL \cite{endo2019animating}  & 38.65  & 0.961  & 62.59  & 0.058 & 441.35 \\
         Deeplandscape \cite{logacheva2020deeplandscape}  & 43.32  & 0.981  & 61.91  & 0.038 & 411.08 \\
         Ours w/o LM  & 36.38  &  0.943 & 63.68  & 0.053 & 830.09  \\
         Ours w/o flip  & 45.99  &  0.981 & 61.82  & 0.025 & 367.77  \\
         Ours w/ $\mathcal{L}_{flow}$ & 41.97  &  0.958 & 61.73  & 0.034 & 365.62 \\
         Ours w/ $\mathcal{L}_{frame}$ & \textbf{46.94}  & \textbf{0.989}  & 61.98  & 0.029 & 370.56 \\
         Ours w/o $\mathcal{L}_{tv}$ & 46.02  & 0.983  & 61.75  & 0.028 & 361.57 \\
         Ours &  46.85 & 0.986  &  \textbf{61.73} &  \textbf{0.025} & \textbf{354.60} \\
         \bottomrule
    \end{tabular}
    \label{tab:comparison2}
    \vspace{-4mm}
\end{threeparttable}
\end{table*}

We conduct a series of quantitative, qualitative experiments and user study to verify the quality of generated videos by our model comparing with existing methods. Three recent DL-based models with released codes are used for comparison: MDGAN \cite{xiong2018learning}, DTVNet \cite{zhang2020dtvnet} and Animating-Landscape (AL) \cite{endo2019animating}. We also try to train Deeplandscape \cite{logacheva2020deeplandscape} on our dataset, but the model is hard to converge and shows unsatisfactory quality. They also reported in paper that their model fails on their new video dataset due to the dataset's scale and motion's distribution. Although our dataset is large, the diversity of scenes and motion make it more difficult for a StyleGAN to converge. Besides, lacking of full temporal dynamics damages the visual effect of scenes other than clouds. Thus, we only do partial experiments on Sky-Timelapse dataset. 

All methods use the first frame of videos in the test set as input and output 32 frames as a sequence at $256\times256$ resolution. 
For MDGAN, we directly fine-tune their released pre-trained model on our new dataset since the two-stage generative model is difficult to train. MDGAN lacks of diverse video generation capabilities and only one output is generated for one input. Thus we can only generate one instance for each test video.

For DTVNet, although the motion vector is normalized, the model cannot guarantee that the flow encoder's output following a certain distribution during training. Thus we randomly sample 5 videos to calculate optical flows using FlowNet 2.0 before feeding them into the flow encoder.  

For AL, we use the same way of saving codebook as motion latent in the paper. During evaluation, five latent codes are randomly sampled from the codebook to generate 5 sequences and we take the average evaluation results of these 5 sequences. 

For our model, we randomly sample 5 videos' embeddings and use them separately to generate 5 videos. Evaluation of each metric is the average of these 5 results. \\

\noindent\textbf{Quantitative Evaluation.} We choose PSNR, SSIM \cite{wang2004image}, FID (Fr\'echet Inception Distance) \cite{heusel2017gans} and LPIPS \cite{zhang2018unreasonable} to evaluate both quality and consistency of generated frames in output videos. These metrics are calculated between first frame and generated video frames. FID and LPIPS are respectively used to measure the quality of generated frames and perceptual dissimilarity between each generated frame and the corresponding first frame. For PSNR and SSIM, we measure the static consistency, which means we need a mask to mask out dynamic region pixels. Given $L$ continuous frames, we calculate the mask as follows:
\begin{equation}
    mask(i, j) = \frac{1}{L}\sum_{i\in[0, L)}\|I_{i+1}^{GT}(i,j)-I_i^{GT}(i,j)\|_{1} > t,
\end{equation}
where threshold $t$ is to 2.5 in our experiments. 
However, perfect image quality and static consistency can be achieved by simply not animating anything at all and they cannot reflect the quality of generated videos. Thus we use Fr\'echet Video Distance \cite{unterthiner2018towards} to measure the quality of generated videos, especially on their motion. User studies are further conducted to evaluate their quality from human sense.

We conduct quantitative experiments on two datasets: our Time-lapse-D and Sky Time-lapse \cite{xiong2018learning}. For evaluation on Sky Time-lapse, we directly use official released models of the three methods.
The comparison results are shown in Table \ref{tab:comparison1} and \ref{tab:comparison2}. Our model gains big improvements on LPIPS and FVD compared with other methods. These two metrics are more close to human perception measures. LPIPS is insensitive to pixel displacement and FVD measures the difference between the generated videos and real videos. Our huge improvement in these two metrics demonstrates that our model can generate more realistic videos. By the better performance of AL and our approach, we found that optical flow can generate better frames compared with 3D-CNN-based methods.
This is because it is difficult for the 3D-CNN-based methods to directly generate high-quality frames. Overall, evaluation results indicate that our approach outperforms the other three baselines in all metrics, which illustrates that our model can generate more high-quality and realistic videos than other methods.

\begin{figure*}[t]
\centering
\includegraphics[width=0.8\paperwidth]{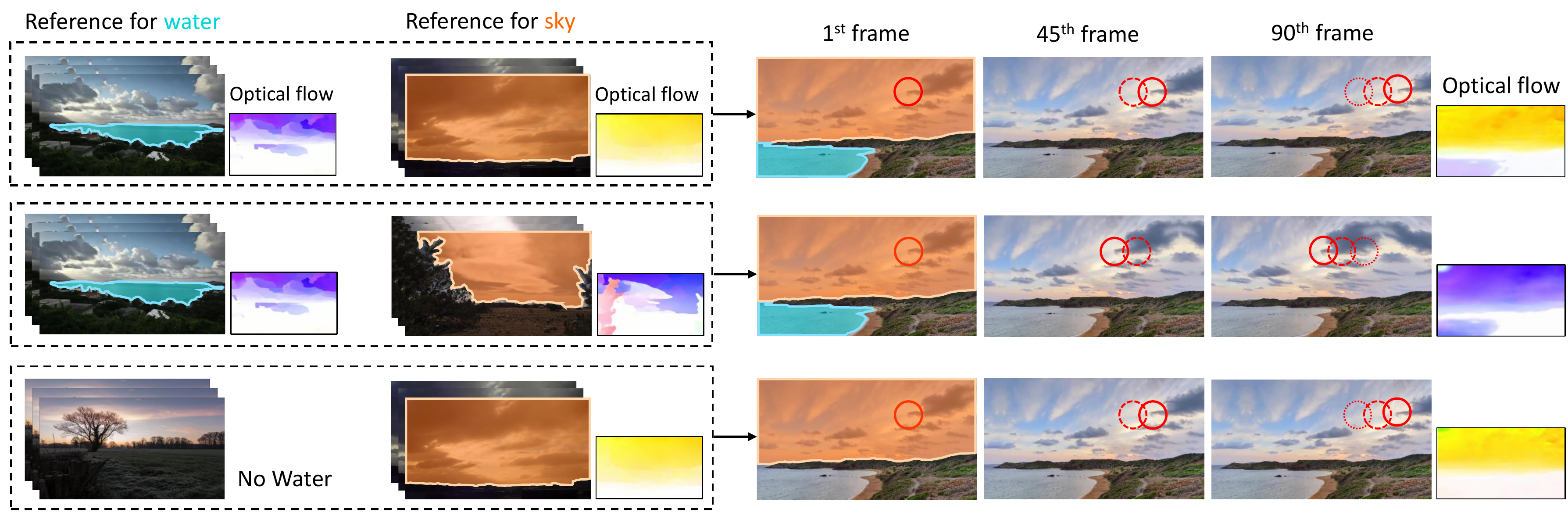}
\vspace{-2mm}
\caption{Visualizations of frames sampled from videos generated with references for water and sky semantics. Different reference videos are used to independently animate different semantics (i.e., sky and water in this figure) of the input image. We sample frames to show the content and visualize optical flows to show the motion. Different colors of visualized optical flow represent different directions and the higher saturation means higher speed and vice versa. [Best viewed in color.]} \label{figure:controllable}
\vspace{-2mm}
\end{figure*}

\begin{figure}[t]
\centering
\includegraphics[width=\linewidth]{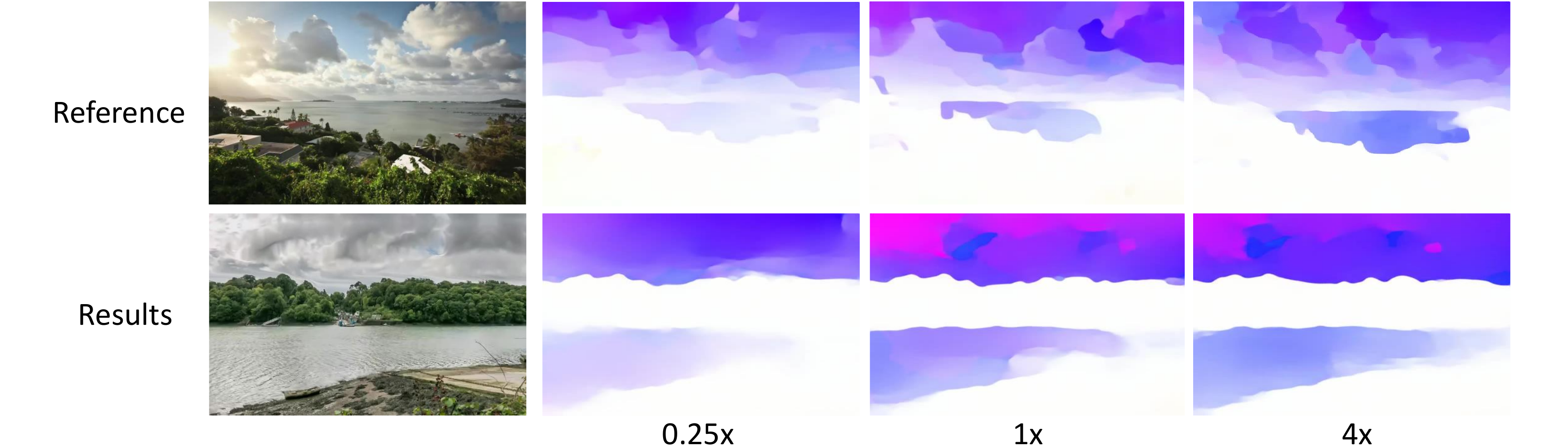}
\vspace{-2mm}
\caption{Sampling from generated videos using reference at different speeds: 0.25x, 1x, 4x. We sample frames to show the content and visualize optical flows to show the motion. Different colors of visualized optical flow represent different directions and the higher saturation means higher speed and vice versa. [Best viewed in color.]}
\label{figure:controllable2}
\end{figure}

\vspace{4mm}
\noindent\textbf{Qualitative Evaluation.}
We give two examples to visually show the result comparison between ours and other approaches in Figure \ref{figure:qualitative}. As in red rectangles that show explicit defects, MDGAN suffers from the distortion in color. DTVNet tends to generate blur frames gradually, which can also be observed from the results shown in their paper. Although AL can generate frames with good details, it often generates distorted motion when multiple semantics with different motion are included in the same frame, which results in unrealistic content distortion in generated frames. Results generated by our method preserve details in high-fidelity while having more obvious motion (as shown by the position of clouds). These results show that our model can generate results with good quality of both content and motion compared with other approaches. 

\noindent\textbf{User Study.} To better evaluate the video quality, we further perform a user study. we randomly choose 100 start images from our test set and generate corresponding videos by different methods using the same settings as described in quantitative evaluation. Then the generated videos by different methods are simultaneously shown to annotators and we can get a result indicating ``which one is better''. Totally, the aforementioned experiment with 700 human subjects is conducted by 14 real workers.  

Comparison results shown in Table \ref{tab:userstudy} demonstrate that our approach outperforms the other three methods by a large margin and our superiority is prominent when comparing with MDGAN and DTVNet. Comparison results with ground-truth videos show that although it is still challenging to generate results that are completely comparable to real videos, our method has the ability to generate some results that can be confused as real videos.

\subsection{Ablation Study}

We design ablation studies to show how each proposal contributes to the final performance. In Table \ref{tab:comparison1} and \ref{tab:comparison2}, w/o LM denotes feeding flow concatenated with segmentation map into a CNN encoder and thus no latent mapping. w/o flip indicates no flipping during training. w/ $\mathcal{L}_{flow}$ and w/ $\mathcal{L}_{frame}$ respectively indicate only use $\mathcal{L}_{flow}$ and $\mathcal{L}_{frame}$ as loss function. w/o $\mathcal{L}_{tv}$ removes the $\mathcal{L}_{tv}$ item compared with our full model. From the results, our full model gains the best performance on most metrics.

\vspace{-2mm}
\subsection{Fine-grained Control of Generation}

In this subsection we use some qualitative results to show the capability of our method to generate videos under fine-grained control. The results are shown in Figure \ref{figure:controllable} and Figure \ref{figure:controllable2}. To facilitate the presentation in the paper, we visualize the optical flow to show the motion of the video. Different colors of visualized optical flow represent different directions and the higher saturation means higher speed and vice versa. As our method can generate videos of any resolution, we unify the resolution of all results to $640 \times 360$. 

In Figure \ref{figure:controllable}, we choose sky and water as two representative semantics in this experiment. From top to down are three animating results from the same input image using different references. First and second row refer to different video references for sky and the same reference for water. According to visualized optical flows, the generated results have similar motion patterns to reference videos on each semantics. First and third row shows that we can controllably fix motion of one semantics by using reference without that semantics. These results demonstrate that our method avoids the global motion problem on multiple semantics. In Figure \ref{figure:controllable2}, we show that our method can also control the speed of motion.

As our approach encodes motion in a fine-grained manner, two benefits are achieved. First, we can controllably animate each semantics in input images to achieve the desired dynamic effect. Secondly, each reference video can be used to extract motion information of one or a few semantics. The constraint of reference videos is greatly reduced, which is obvious while animating an image with many semantics.

\vspace{-3mm}
\section{Conclusions}
In this paper, we propose a fine-grained motion embedding approach to animate single landscape images to time-lapse videos with accurate and realistic motion. Our model consists of two parts: a fine-grained motion encoder independently encodes the motion patterns of different semantics. Then a motion generator generates motion for animation. The result video is generated by warping frames with the generated motion in an accumulated way. We collect the largest high-resolution time-lapse video dataset (Time-lapse-D) with multiple scenes. Experimental results show that our approach can generate more realistic and high quality videos compared to state-of-the-art methods by both quantitative and qualitative evaluations. In the future, we will study on robust approach to deal with more diverse scenes, such as flower blossoming and traffic flow from the view of style transfer \cite{hu2020aesthetic}. And methods used in video inpainting \cite{yan2019PENnet,yan2020sttn} and super resolution \cite{yang2020learning} can also be applied to overcome the occlusion.

\begin{acks}
This work was supported by the National Natural Science Foundation of China (NSFC) under Grants 61836011 and 62021001. The authors would like to thank all users for voluntarily participating our user study.  
\end{acks}

\bibliographystyle{ACM-Reference-Format}
\balance
\bibliography{main}


\begin{thebibliography}{36}


\ifx \showCODEN    \undefined \def \showCODEN     #1{\unskip}     \fi
\ifx \showDOI      \undefined \def \showDOI       #1{#1}\fi
\ifx \showISBNx    \undefined \def \showISBNx     #1{\unskip}     \fi
\ifx \showISBNxiii \undefined \def \showISBNxiii  #1{\unskip}     \fi
\ifx \showISSN     \undefined \def \showISSN      #1{\unskip}     \fi
\ifx \showLCCN     \undefined \def \showLCCN      #1{\unskip}     \fi
\ifx \shownote     \undefined \def \shownote      #1{#1}          \fi
\ifx \showarticletitle \undefined \def \showarticletitle #1{#1}   \fi
\ifx \showURL      \undefined \def \showURL       {\relax}        \fi
\providecommand\bibfield[2]{#2}
\providecommand\bibinfo[2]{#2}
\providecommand\natexlab[1]{#1}
\providecommand\showeprint[2][]{arXiv:#2}

\bibitem[\protect\citeauthoryear{Cheng, Chen, and Chiu}{Cheng
  et~al\mbox{.}}{2020}]%
        {cheng2020time}
\bibfield{author}{\bibinfo{person}{Chia-Chi Cheng}, \bibinfo{person}{Hung-Yu
  Chen}, {and} \bibinfo{person}{Wei-Chen Chiu}.}
  \bibinfo{year}{2020}\natexlab{}.
\newblock \showarticletitle{Time Flies: Animating a Still Image With Time-Lapse
  Video As Reference}. In \bibinfo{booktitle}{\emph{CVPR}}.
  \bibinfo{pages}{5641--5650}.
\newblock


\bibitem[\protect\citeauthoryear{Endo, Kanamori, and Kuriyama}{Endo
  et~al\mbox{.}}{2019}]%
        {endo2019animating}
\bibfield{author}{\bibinfo{person}{Yuki Endo}, \bibinfo{person}{Yoshihiro
  Kanamori}, {and} \bibinfo{person}{Shigeru Kuriyama}.}
  \bibinfo{year}{2019}\natexlab{}.
\newblock \showarticletitle{Animating landscape: self-supervised learning of
  decoupled motion and appearance for single-image video synthesis}.
\newblock \bibinfo{journal}{\emph{TOG}} \bibinfo{volume}{38},
  \bibinfo{number}{6} (\bibinfo{year}{2019}), \bibinfo{pages}{1--19}.
\newblock


\bibitem[\protect\citeauthoryear{Goodfellow, Pouget-Abadie, Mirza, Xu,
  Warde-Farley, Ozair, Courville, and Bengio}{Goodfellow et~al\mbox{.}}{2014}]%
        {goodfellow2014generative}
\bibfield{author}{\bibinfo{person}{Ian Goodfellow}, \bibinfo{person}{Jean
  Pouget-Abadie}, \bibinfo{person}{Mehdi Mirza}, \bibinfo{person}{Bing Xu},
  \bibinfo{person}{David Warde-Farley}, \bibinfo{person}{Sherjil Ozair},
  \bibinfo{person}{Aaron Courville}, {and} \bibinfo{person}{Yoshua Bengio}.}
  \bibinfo{year}{2014}\natexlab{}.
\newblock \showarticletitle{Generative adversarial nets}. In
  \bibinfo{booktitle}{\emph{NeurIPS}}. \bibinfo{pages}{2672--2680}.
\newblock


\bibitem[\protect\citeauthoryear{Hao, Huang, and Belongie}{Hao
  et~al\mbox{.}}{2018}]%
        {hao2018controllable}
\bibfield{author}{\bibinfo{person}{Zekun Hao}, \bibinfo{person}{Xun Huang},
  {and} \bibinfo{person}{Serge Belongie}.} \bibinfo{year}{2018}\natexlab{}.
\newblock \showarticletitle{Controllable video generation with sparse
  trajectories}. In \bibinfo{booktitle}{\emph{CVPR}}.
  \bibinfo{pages}{7854--7863}.
\newblock


\bibitem[\protect\citeauthoryear{Heusel, Ramsauer, Unterthiner, Nessler, and
  Hochreiter}{Heusel et~al\mbox{.}}{2017}]%
        {heusel2017gans}
\bibfield{author}{\bibinfo{person}{Martin Heusel}, \bibinfo{person}{Hubert
  Ramsauer}, \bibinfo{person}{Thomas Unterthiner}, \bibinfo{person}{Bernhard
  Nessler}, {and} \bibinfo{person}{Sepp Hochreiter}.}
  \bibinfo{year}{2017}\natexlab{}.
\newblock \showarticletitle{{GANs} trained by a two time-scale update rule
  converge to a local nash equilibrium}. In
  \bibinfo{booktitle}{\emph{NeurIPS}}. \bibinfo{pages}{6626--6637}.
\newblock


\bibitem[\protect\citeauthoryear{Hu, Waelchli, Portenier, Zwicker, and
  Favaro}{Hu et~al\mbox{.}}{2018}]%
        {hu2018video}
\bibfield{author}{\bibinfo{person}{Qiyang Hu}, \bibinfo{person}{Adrian
  Waelchli}, \bibinfo{person}{Tiziano Portenier}, \bibinfo{person}{Matthias
  Zwicker}, {and} \bibinfo{person}{Paolo Favaro}.}
  \bibinfo{year}{2018}\natexlab{}.
\newblock \showarticletitle{Video synthesis from a single image and motion
  stroke}.
\newblock \bibinfo{journal}{\emph{arXiv preprint arXiv:1812.01874}}
  (\bibinfo{year}{2018}).
\newblock


\bibitem[\protect\citeauthoryear{Hu, Jia, Liu, Bu, and Fu}{Hu
  et~al\mbox{.}}{2020}]%
        {hu2020aesthetic}
\bibfield{author}{\bibinfo{person}{Zhiyuan Hu}, \bibinfo{person}{Jia Jia},
  \bibinfo{person}{Bei Liu}, \bibinfo{person}{Yaohua Bu}, {and}
  \bibinfo{person}{Jianlong Fu}.} \bibinfo{year}{2020}\natexlab{}.
\newblock \showarticletitle{Aesthetic-aware image style transfer}. In
  \bibinfo{booktitle}{\emph{Proceedings of the 28th ACM International
  Conference on Multimedia}}. \bibinfo{pages}{3320--3329}.
\newblock


\bibitem[\protect\citeauthoryear{Ilg, Mayer, Saikia, Keuper, Dosovitskiy, and
  Brox}{Ilg et~al\mbox{.}}{2017}]%
        {ilg2017flownet}
\bibfield{author}{\bibinfo{person}{Eddy Ilg}, \bibinfo{person}{Nikolaus Mayer},
  \bibinfo{person}{Tonmoy Saikia}, \bibinfo{person}{Margret Keuper},
  \bibinfo{person}{Alexey Dosovitskiy}, {and} \bibinfo{person}{Thomas Brox}.}
  \bibinfo{year}{2017}\natexlab{}.
\newblock \showarticletitle{Flownet 2.0: Evolution of optical flow estimation
  with deep networks}. In \bibinfo{booktitle}{\emph{CVPR}}.
  \bibinfo{pages}{2462--2470}.
\newblock


\bibitem[\protect\citeauthoryear{Isola, Zhu, Zhou, and Efros}{Isola
  et~al\mbox{.}}{2017}]%
        {isola2017image}
\bibfield{author}{\bibinfo{person}{Phillip Isola}, \bibinfo{person}{Jun-Yan
  Zhu}, \bibinfo{person}{Tinghui Zhou}, {and} \bibinfo{person}{Alexei~A
  Efros}.} \bibinfo{year}{2017}\natexlab{}.
\newblock \showarticletitle{Image-to-image translation with conditional
  adversarial networks}. In \bibinfo{booktitle}{\emph{CVPR}}.
  \bibinfo{pages}{1125--1134}.
\newblock


\bibitem[\protect\citeauthoryear{Karras, Laine, and Aila}{Karras
  et~al\mbox{.}}{2019}]%
        {karras2019style}
\bibfield{author}{\bibinfo{person}{Tero Karras}, \bibinfo{person}{Samuli
  Laine}, {and} \bibinfo{person}{Timo Aila}.} \bibinfo{year}{2019}\natexlab{}.
\newblock \showarticletitle{A style-based generator architecture for generative
  adversarial networks}. In \bibinfo{booktitle}{\emph{CVPR}}.
  \bibinfo{pages}{4401--4410}.
\newblock


\bibitem[\protect\citeauthoryear{Kingma and Ba}{Kingma and Ba}{2014}]%
        {kingma2014adam}
\bibfield{author}{\bibinfo{person}{Diederik~P Kingma} {and}
  \bibinfo{person}{Jimmy Ba}.} \bibinfo{year}{2014}\natexlab{}.
\newblock \showarticletitle{Adam: A method for stochastic optimization}.
\newblock \bibinfo{journal}{\emph{arXiv preprint arXiv:1412.6980}}
  (\bibinfo{year}{2014}).
\newblock


\bibitem[\protect\citeauthoryear{Li, Fang, Yang, Wang, Lu, and Yang}{Li
  et~al\mbox{.}}{2018}]%
        {li2018flow}
\bibfield{author}{\bibinfo{person}{Yijun Li}, \bibinfo{person}{Chen Fang},
  \bibinfo{person}{Jimei Yang}, \bibinfo{person}{Zhaowen Wang},
  \bibinfo{person}{Xin Lu}, {and} \bibinfo{person}{Ming-Hsuan Yang}.}
  \bibinfo{year}{2018}\natexlab{}.
\newblock \showarticletitle{Flow-grounded spatial-temporal video prediction
  from still images}. In \bibinfo{booktitle}{\emph{ECCV}}.
  \bibinfo{pages}{600--615}.
\newblock


\bibitem[\protect\citeauthoryear{Liu, Reda, Shih, Wang, Tao, and Catanzaro}{Liu
  et~al\mbox{.}}{2018}]%
        {liu2018image}
\bibfield{author}{\bibinfo{person}{Guilin Liu}, \bibinfo{person}{Fitsum~A
  Reda}, \bibinfo{person}{Kevin~J Shih}, \bibinfo{person}{Ting-Chun Wang},
  \bibinfo{person}{Andrew Tao}, {and} \bibinfo{person}{Bryan Catanzaro}.}
  \bibinfo{year}{2018}\natexlab{}.
\newblock \showarticletitle{Image inpainting for irregular holes using partial
  convolutions}. In \bibinfo{booktitle}{\emph{ECCV}}. \bibinfo{pages}{85--100}.
\newblock


\bibitem[\protect\citeauthoryear{Logacheva, Suvorov, Khomenko, Mashikhin, and
  Lempitsky}{Logacheva et~al\mbox{.}}{2020}]%
        {logacheva2020deeplandscape}
\bibfield{author}{\bibinfo{person}{Elizaveta Logacheva}, \bibinfo{person}{Roman
  Suvorov}, \bibinfo{person}{Oleg Khomenko}, \bibinfo{person}{Anton Mashikhin},
  {and} \bibinfo{person}{Victor Lempitsky}.} \bibinfo{year}{2020}\natexlab{}.
\newblock \showarticletitle{DeepLandscape: adversarial modeling of landscape
  video}. In \bibinfo{booktitle}{\emph{ECCV}}. \bibinfo{pages}{256--272}.
\newblock


\bibitem[\protect\citeauthoryear{Nam, Ma, Chai, Brendel, Xu, and Kim}{Nam
  et~al\mbox{.}}{2019}]%
        {nam2019end}
\bibfield{author}{\bibinfo{person}{Seonghyeon Nam}, \bibinfo{person}{Chongyang
  Ma}, \bibinfo{person}{Menglei Chai}, \bibinfo{person}{William Brendel},
  \bibinfo{person}{Ning Xu}, {and} \bibinfo{person}{Seon~Joo Kim}.}
  \bibinfo{year}{2019}\natexlab{}.
\newblock \showarticletitle{End-to-end time-lapse video synthesis from a single
  outdoor image}. In \bibinfo{booktitle}{\emph{CVPR}}.
  \bibinfo{pages}{1409--1418}.
\newblock


\bibitem[\protect\citeauthoryear{Okabe, Anjyo, Igarashi, and Seidel}{Okabe
  et~al\mbox{.}}{2009}]%
        {okabe2009animating}
\bibfield{author}{\bibinfo{person}{Makoto Okabe}, \bibinfo{person}{Ken Anjyo},
  \bibinfo{person}{Takeo Igarashi}, {and} \bibinfo{person}{Hans-Peter Seidel}.}
  \bibinfo{year}{2009}\natexlab{}.
\newblock \showarticletitle{Animating pictures of fluid using video examples}.
  In \bibinfo{booktitle}{\emph{CGF}}, Vol.~\bibinfo{volume}{28}.
  \bibinfo{pages}{677--686}.
\newblock


\bibitem[\protect\citeauthoryear{Okabe, Anjyor, and Onai}{Okabe
  et~al\mbox{.}}{2011}]%
        {okabe2011creating}
\bibfield{author}{\bibinfo{person}{Makoto Okabe}, \bibinfo{person}{Ken Anjyor},
  {and} \bibinfo{person}{Rikio Onai}.} \bibinfo{year}{2011}\natexlab{}.
\newblock \showarticletitle{Creating fluid animation from a single image using
  video database}. In \bibinfo{booktitle}{\emph{CGF}},
  Vol.~\bibinfo{volume}{30}. \bibinfo{pages}{1973--1982}.
\newblock


\bibitem[\protect\citeauthoryear{Okabe, Dobashi, and Anjyo}{Okabe
  et~al\mbox{.}}{2018}]%
        {okabe2018animating}
\bibfield{author}{\bibinfo{person}{Makoto Okabe}, \bibinfo{person}{Yoshinori
  Dobashi}, {and} \bibinfo{person}{Ken Anjyo}.}
  \bibinfo{year}{2018}\natexlab{}.
\newblock \showarticletitle{Animating pictures of water scenes using video
  retrieval}.
\newblock \bibinfo{journal}{\emph{The Visual Computer}} \bibinfo{volume}{34},
  \bibinfo{number}{3} (\bibinfo{year}{2018}), \bibinfo{pages}{347--358}.
\newblock


\bibitem[\protect\citeauthoryear{Pan, Wang, Jia, Shao, Sheng, Yan, and
  Wang}{Pan et~al\mbox{.}}{2019}]%
        {pan2019video}
\bibfield{author}{\bibinfo{person}{Junting Pan}, \bibinfo{person}{Chengyu
  Wang}, \bibinfo{person}{Xu Jia}, \bibinfo{person}{Jing Shao},
  \bibinfo{person}{Lu Sheng}, \bibinfo{person}{Junjie Yan}, {and}
  \bibinfo{person}{Xiaogang Wang}.} \bibinfo{year}{2019}\natexlab{}.
\newblock \showarticletitle{Video generation from single semantic label map}.
  In \bibinfo{booktitle}{\emph{CVPR}}. \bibinfo{pages}{3733--3742}.
\newblock


\bibitem[\protect\citeauthoryear{Prashnani, Noorkami, Vaquero, and
  Sen}{Prashnani et~al\mbox{.}}{2017}]%
        {prashnani2017phase}
\bibfield{author}{\bibinfo{person}{Ekta Prashnani}, \bibinfo{person}{Maneli
  Noorkami}, \bibinfo{person}{Daniel Vaquero}, {and} \bibinfo{person}{Pradeep
  Sen}.} \bibinfo{year}{2017}\natexlab{}.
\newblock \showarticletitle{A phase-based approach for animating images using
  video examples}. In \bibinfo{booktitle}{\emph{CGF}},
  Vol.~\bibinfo{volume}{36}. \bibinfo{pages}{303--311}.
\newblock


\bibitem[\protect\citeauthoryear{Shih, Paris, Durand, and Freeman}{Shih
  et~al\mbox{.}}{2013}]%
        {shih2013data}
\bibfield{author}{\bibinfo{person}{Yichang Shih}, \bibinfo{person}{Sylvain
  Paris}, \bibinfo{person}{Fr{\'e}do Durand}, {and} \bibinfo{person}{William~T
  Freeman}.} \bibinfo{year}{2013}\natexlab{}.
\newblock \showarticletitle{Data-driven hallucination of different times of day
  from a single outdoor photo}.
\newblock \bibinfo{journal}{\emph{TOG}} \bibinfo{volume}{32},
  \bibinfo{number}{6} (\bibinfo{year}{2013}), \bibinfo{pages}{1--11}.
\newblock


\bibitem[\protect\citeauthoryear{Sun, Zhao, Jiang, Cheng, Xiao, Liu, Mu, Wang,
  Liu, and Wang}{Sun et~al\mbox{.}}{2019}]%
        {sun2019high}
\bibfield{author}{\bibinfo{person}{Ke Sun}, \bibinfo{person}{Yang Zhao},
  \bibinfo{person}{Borui Jiang}, \bibinfo{person}{Tianheng Cheng},
  \bibinfo{person}{Bin Xiao}, \bibinfo{person}{Dong Liu},
  \bibinfo{person}{Yadong Mu}, \bibinfo{person}{Xinggang Wang},
  \bibinfo{person}{Wenyu Liu}, {and} \bibinfo{person}{Jingdong Wang}.}
  \bibinfo{year}{2019}\natexlab{}.
\newblock \showarticletitle{High-resolution representations for labeling pixels
  and regions}.
\newblock \bibinfo{journal}{\emph{arXiv preprint arXiv:1904.04514}}
  (\bibinfo{year}{2019}).
\newblock


\bibitem[\protect\citeauthoryear{Tulyakov, Liu, Yang, and Kautz}{Tulyakov
  et~al\mbox{.}}{2018}]%
        {tulyakov2018mocogan}
\bibfield{author}{\bibinfo{person}{Sergey Tulyakov}, \bibinfo{person}{Ming-Yu
  Liu}, \bibinfo{person}{Xiaodong Yang}, {and} \bibinfo{person}{Jan Kautz}.}
  \bibinfo{year}{2018}\natexlab{}.
\newblock \showarticletitle{Mocogan: Decomposing motion and content for video
  generation}. In \bibinfo{booktitle}{\emph{CVPR}}.
  \bibinfo{pages}{1526--1535}.
\newblock


\bibitem[\protect\citeauthoryear{Ulyanov, Vedaldi, and Lempitsky}{Ulyanov
  et~al\mbox{.}}{2016}]%
        {ulyanov2016instance}
\bibfield{author}{\bibinfo{person}{Dmitry Ulyanov}, \bibinfo{person}{Andrea
  Vedaldi}, {and} \bibinfo{person}{Victor Lempitsky}.}
  \bibinfo{year}{2016}\natexlab{}.
\newblock \showarticletitle{Instance normalization: The missing ingredient for
  fast stylization}.
\newblock \bibinfo{journal}{\emph{arXiv preprint arXiv:1607.08022}}
  (\bibinfo{year}{2016}).
\newblock


\bibitem[\protect\citeauthoryear{Unterthiner, van Steenkiste, Kurach, Marinier,
  Michalski, and Gelly}{Unterthiner et~al\mbox{.}}{2018}]%
        {unterthiner2018towards}
\bibfield{author}{\bibinfo{person}{Thomas Unterthiner}, \bibinfo{person}{Sjoerd
  van Steenkiste}, \bibinfo{person}{Karol Kurach}, \bibinfo{person}{Raphael
  Marinier}, \bibinfo{person}{Marcin Michalski}, {and} \bibinfo{person}{Sylvain
  Gelly}.} \bibinfo{year}{2018}\natexlab{}.
\newblock \showarticletitle{Towards accurate generative models of video: A new
  metric \& challenges}.
\newblock \bibinfo{journal}{\emph{arXiv preprint arXiv:1812.01717}}
  (\bibinfo{year}{2018}).
\newblock


\bibitem[\protect\citeauthoryear{Wang, Bovik, Sheikh, and Simoncelli}{Wang
  et~al\mbox{.}}{2004}]%
        {wang2004image}
\bibfield{author}{\bibinfo{person}{Zhou Wang}, \bibinfo{person}{Alan~C Bovik},
  \bibinfo{person}{Hamid~R Sheikh}, {and} \bibinfo{person}{Eero~P Simoncelli}.}
  \bibinfo{year}{2004}\natexlab{}.
\newblock \showarticletitle{Image quality assessment: from error visibility to
  structural similarity}.
\newblock \bibinfo{journal}{\emph{TIP}} \bibinfo{volume}{13},
  \bibinfo{number}{4} (\bibinfo{year}{2004}), \bibinfo{pages}{600--612}.
\newblock


\bibitem[\protect\citeauthoryear{Xiong, Luo, Ma, Liu, and Luo}{Xiong
  et~al\mbox{.}}{2018}]%
        {xiong2018learning}
\bibfield{author}{\bibinfo{person}{Wei Xiong}, \bibinfo{person}{Wenhan Luo},
  \bibinfo{person}{Lin Ma}, \bibinfo{person}{Wei Liu}, {and}
  \bibinfo{person}{Jiebo Luo}.} \bibinfo{year}{2018}\natexlab{}.
\newblock \showarticletitle{Learning to generate time-lapse videos using
  multi-stage dynamic generative adversarial networks}. In
  \bibinfo{booktitle}{\emph{CVPR}}. \bibinfo{pages}{2364--2373}.
\newblock


\bibitem[\protect\citeauthoryear{Xu, Wang, Chen, and Li}{Xu
  et~al\mbox{.}}{2015}]%
        {xu2015empirical}
\bibfield{author}{\bibinfo{person}{Bing Xu}, \bibinfo{person}{Naiyan Wang},
  \bibinfo{person}{Tianqi Chen}, {and} \bibinfo{person}{Mu Li}.}
  \bibinfo{year}{2015}\natexlab{}.
\newblock \showarticletitle{Empirical evaluation of rectified activations in
  convolutional network}.
\newblock \bibinfo{journal}{\emph{arXiv preprint arXiv:1505.00853}}
  (\bibinfo{year}{2015}).
\newblock


\bibitem[\protect\citeauthoryear{Yang, Yang, Fu, Lu, and Guo}{Yang
  et~al\mbox{.}}{2020}]%
        {yang2020learning}
\bibfield{author}{\bibinfo{person}{Fuzhi Yang}, \bibinfo{person}{Huan Yang},
  \bibinfo{person}{Jianlong Fu}, \bibinfo{person}{Hongtao Lu}, {and}
  \bibinfo{person}{Baining Guo}.} \bibinfo{year}{2020}\natexlab{}.
\newblock \showarticletitle{Learning Texture Transformer Network for Image
  Super-Resolution}. In \bibinfo{booktitle}{\emph{CVPR}}.
\newblock


\bibitem[\protect\citeauthoryear{Zeng, Fu, and Chao}{Zeng
  et~al\mbox{.}}{2020}]%
        {yan2020sttn}
\bibfield{author}{\bibinfo{person}{Yanhong Zeng}, \bibinfo{person}{Jianlong
  Fu}, {and} \bibinfo{person}{Hongyang Chao}.} \bibinfo{year}{2020}\natexlab{}.
\newblock \showarticletitle{Learning Joint Spatial-Temporal Transformations for
  Video Inpainting}. In \bibinfo{booktitle}{\emph{ECCV}}.
\newblock


\bibitem[\protect\citeauthoryear{Zeng, Fu, Chao, and Guo}{Zeng
  et~al\mbox{.}}{2019}]%
        {yan2019PENnet}
\bibfield{author}{\bibinfo{person}{Yanhong Zeng}, \bibinfo{person}{Jianlong
  Fu}, \bibinfo{person}{Hongyang Chao}, {and} \bibinfo{person}{Baining Guo}.}
  \bibinfo{year}{2019}\natexlab{}.
\newblock \showarticletitle{Learning Pyramid-Context Encoder Network for
  High-Quality Image Inpainting}. In \bibinfo{booktitle}{\emph{CVPR}}.
  \bibinfo{pages}{1486--1494}.
\newblock


\bibitem[\protect\citeauthoryear{Zhang, Xu, Liu, Wang, Wu, Liu, and
  Jiang}{Zhang et~al\mbox{.}}{2020}]%
        {zhang2020dtvnet}
\bibfield{author}{\bibinfo{person}{Jiangning Zhang}, \bibinfo{person}{Chao Xu},
  \bibinfo{person}{Liang Liu}, \bibinfo{person}{Mengmeng Wang},
  \bibinfo{person}{Xia Wu}, \bibinfo{person}{Yong Liu}, {and}
  \bibinfo{person}{Yunliang Jiang}.} \bibinfo{year}{2020}\natexlab{}.
\newblock \showarticletitle{{DTVNet}: Dynamic Time-lapse Video Generation via
  Single Still Image}. In \bibinfo{booktitle}{\emph{ECCV}}.
  \bibinfo{pages}{300--315}.
\newblock


\bibitem[\protect\citeauthoryear{Zhang, Isola, Efros, Shechtman, and
  Wang}{Zhang et~al\mbox{.}}{2018}]%
        {zhang2018unreasonable}
\bibfield{author}{\bibinfo{person}{Richard Zhang}, \bibinfo{person}{Phillip
  Isola}, \bibinfo{person}{Alexei~A Efros}, \bibinfo{person}{Eli Shechtman},
  {and} \bibinfo{person}{Oliver Wang}.} \bibinfo{year}{2018}\natexlab{}.
\newblock \showarticletitle{The unreasonable effectiveness of deep features as
  a perceptual metric}. In \bibinfo{booktitle}{\emph{CVPR}}.
  \bibinfo{pages}{586--595}.
\newblock


\bibitem[\protect\citeauthoryear{Zhao, Peng, Tian, Kapadia, and Metaxas}{Zhao
  et~al\mbox{.}}{2018}]%
        {zhao2018learning}
\bibfield{author}{\bibinfo{person}{Long Zhao}, \bibinfo{person}{Xi Peng},
  \bibinfo{person}{Yu Tian}, \bibinfo{person}{Mubbasir Kapadia}, {and}
  \bibinfo{person}{Dimitris Metaxas}.} \bibinfo{year}{2018}\natexlab{}.
\newblock \showarticletitle{Learning to forecast and refine residual motion for
  image-to-video generation}. In \bibinfo{booktitle}{\emph{ECCV}}.
  \bibinfo{pages}{387--403}.
\newblock


\bibitem[\protect\citeauthoryear{Zhou, Zhao, Puig, Xiao, Fidler, Barriuso, and
  Torralba}{Zhou et~al\mbox{.}}{2019}]%
        {zhou2019semantic}
\bibfield{author}{\bibinfo{person}{Bolei Zhou}, \bibinfo{person}{Hang Zhao},
  \bibinfo{person}{Xavier Puig}, \bibinfo{person}{Tete Xiao},
  \bibinfo{person}{Sanja Fidler}, \bibinfo{person}{Adela Barriuso}, {and}
  \bibinfo{person}{Antonio Torralba}.} \bibinfo{year}{2019}\natexlab{}.
\newblock \showarticletitle{Semantic understanding of scenes through the
  {ADE20K} dataset}.
\newblock \bibinfo{journal}{\emph{IJCV}} \bibinfo{volume}{127},
  \bibinfo{number}{3} (\bibinfo{year}{2019}), \bibinfo{pages}{302--321}.
\newblock


\bibitem[\protect\citeauthoryear{Zhou, Song, and Berg}{Zhou
  et~al\mbox{.}}{2018}]%
        {zhou2018image2gif}
\bibfield{author}{\bibinfo{person}{Yipin Zhou}, \bibinfo{person}{Yale Song},
  {and} \bibinfo{person}{Tamara~L Berg}.} \bibinfo{year}{2018}\natexlab{}.
\newblock \showarticletitle{Image2GIF: Generating cinemagraphs using recurrent
  deep {Q-Networks}}. In \bibinfo{booktitle}{\emph{WACV}}.
  \bibinfo{pages}{170--178}.
\newblock


\end{thebibliography}


\end{document}